# MALTopic: Multi-Agent LLM Topic Modeling Framework

Yash Sharma
*Independent Researcher*
San Francisco, USA
yash91sharma@gmail.com

*Abstract*—Topic modeling is a crucial technique for extracting latent themes from unstructured text data, particularly valuable in analyzing survey responses. However, traditional methods often only consider free-text responses and do not natively incorporate structured or categorical survey responses for topic modeling. And they produce abstract topics, requiring extensive human interpretation. To address these limitations, we propose the Multi-Agent LLM Topic Modeling Framework (MALTopic). This framework decomposes topic modeling into specialized tasks executed by individual LLM agents: an enrichment agent leverages structured data to enhance textual responses, a topic modeling agent extracts latent themes, and a deduplication agent refines the results. Comparative analysis on a survey dataset demonstrates that MALTopic significantly improves topic coherence, diversity, and interpretability compared to LDA and BERTopic. By integrating structured data and employing a multi-agent approach, MALTopic generates human-readable topics with enhanced contextual relevance, offering a more effective solution for analyzing complex survey data.

## I. INTRODUCTION

Topic modeling is a widely-used natural language processing (NLP) technique for extracting latent thematic structures from unstructured text data. It can be very helpful in analyzing survey responses and understanding various aspects of human experience.

However, traditional topic modeling algorithms (like LDA [1], BERTopic [2]) have two major challenges when it comes to topic modeling on survey data.

- First, they primarily focus on analyzing free-text (open-ended text) data in isolation, and overlook the rich context provided by structured or categorical data.
- Second, while these algorithms can automatically identify latent themes, the resulting topics are often abstract. And require significant human intervention for interpretation of these topics.

To address these limitations, we propose the Multi-Agent LLM Topic Modeling Framework (MALTopic). We decompose topic modeling into smaller independent tasks, which are executed by individual agents using LLM. First, we propose an **enrichment agent** that leverages structured and categorical survey responses to enhance the semantic context of free-text responses. Second, a **topic modeling agent** extracts latent themes from the enriched responses, followed by a **topic deduplication agent** to refine and consolidate the identified topics. This multi-agent approach aims to natively consume structured and categorical data along with free-text responses and generate a set of human-readable topics.

For comparison, we performed topic modeling on a survey response dataset [3] using MALTopic, LDA and BERTopic. The results were then compared using quantitative and qualitative analysis.

## II. RELATED WORK

The limitations of topic modeling techniques have spurred recent investigations into the application of LLMs [4]-[8]. Notably, researchers have explored LLM based topic modeling for specific challenges, including short texts [9]-[11] and small datasets [12]. However, a significant research gap persists: the utilization of LLMs for topic modeling on survey data comprising both free-text and structured responses remains unexplored.

The concept of multi-stage architectures in topic modeling is not entirely new. Frameworks like BERTopic [2] employ a modular design that allows for a sequence of steps, like embedding generation, dimensionality reduction, clustering, and topic representation. The idea of using multiple specialized agents, particularly LLMs, to tackle different facets of a complex task aligns with the broader trend of multi-agent AI systems [13]. The MALTopic framework, with its decomposition of topic modeling into distinct tasks handled by specialized LLM agents, builds upon these existing concepts of multi-stage processing and multi-agent collaboration.

The abbreviation for the proposed framework, MALTopic, draws inspiration from BERTopic [2], a topic modeling technique that leverages transformers and the class-based term frequency-inverse document frequency (c-TF-IDF).

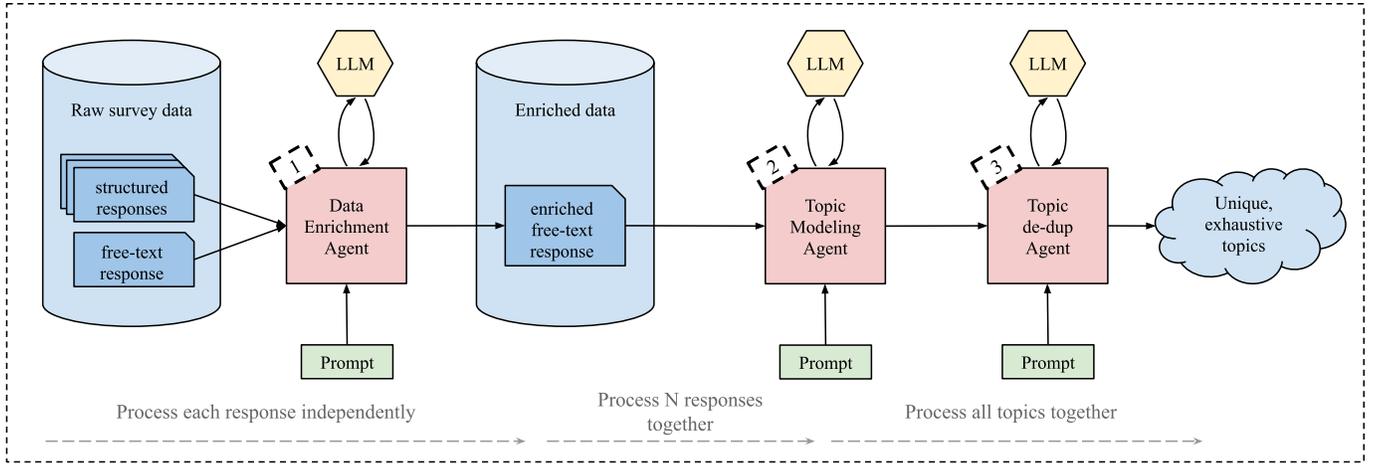

Fig. 1. MALTopic Framework Overview

## III. METHODOLOGY

### A. Survey Dataset

This research leverages survey response dataset for topic modeling. This anonymous survey was conducted on 202 professionals to understand the impact of AI tools on their work. The respondents were from engineering, data science, analytics, leadership, product management and project management fields. Full survey questions with example responses are available online [3].

The survey incorporated six structured/categorical questions. This included job title, work experience, day-to-day tasks, major challenges at work and opinions on the applicability of AI tools to specific tasks.

Furthermore, two free-text questions were included to elicit qualitative data:

- AI tools in tech: What are your biggest concerns?
- AI tools in tech: What excites you the most?

These free-text responses were enriched with structured questions and used for topic modeling, as explained below.

This survey was chosen because of its appropriate size(N=202), enabling human validations and oversight on the topic modeling process for this research.

### B. MALTopic System Overview

Fig. 1 explains the MALTopic framework architecture. There are three agents, each operates independently with a very specific purpose.

**Agent 1 - Data Enrichment Agent** employs an LLM to enrich the free-text response with the structured responses. LLM's powerful natural language capabilities produce an enriched semantic representation of each survey response, embedding both linguistic and respondent-specific context.

Each free-text response undergoes independent enrichment using the structured data. So for an input of N responses, the output is N enriched responses.

For this research, we performed topic modeling on both the free-text responses (concerns, and excitement about AI tools in tech) separately. Both were enriched with two structured responses: job title and work experience. Fig. 2 shows an example of enrichment by Agent 1.

**Agent 2 - Topic Modeling Agent** uses an LLM to find topics in the enriched responses. By using the enriched responses, the LLM can capture complex semantic nuances, interdependencies and latent themes, leading to more precise and context aware topic extraction.

The responses are batch processed. Which means the LLM concurrently analyzes a batch of N enriched responses, constrained by the model's input token limit, and yields M exhaustive, non-overlapping, and distinct topics.

**Agent 3 - Topic Deduplication Agent** performs a deduplication of the topic list provided by Agent 2. This is required in scenarios involving a large number of survey responses that exceed Agent 2's capacity for single-pass processing.

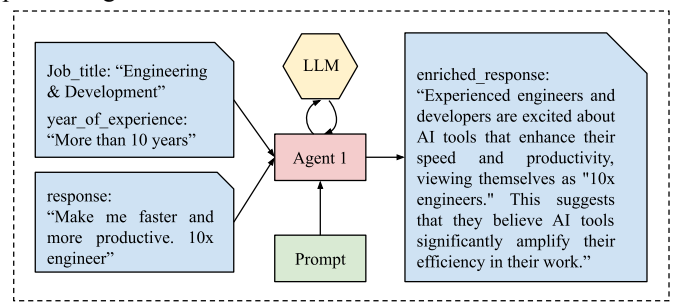

Fig. 2. Example enrichment by Agent 1

## C. LLM Model

For the MALTopic research, we employed the GPT-4o-mini-2024-07-18 language model [14]. This model was configured with following custom parameter values:

- seed: 1234
- temperature: 0.2
- top_p: 0.9

All other model parameters were maintained at their default settings. This model supports 128K input tokens, and 16K output tokens. At the time of this research, the cost of using the model via OpenAI API [15] was USD \$0.15 per million tokens of input, and USD \$0.075 per million tokens of output.

## D. Prompts

Each agent employed a single-shot prompting strategy. As an illustration, the LLM prompt for data enrichment (Agent 1) is shown in Fig. 3.

Due to length constraints, the prompts utilized across all agents are not included herein. However, the complete text of the prompts employed with each MALTopic agent is available online [3]. It is important to note that the prompt for topic modeling (Agent 2) explicitly instructed the LLM to identify unique, non-overlapping and exhaustive topics, and for each topic generate the following:

- Topic Name: A concise and descriptive name for the topic
- Description: A one line summary of what the topic encompasses.
- Respondent Profile: Which respondent profiles are particularly relevant to this topic
- Representative words: List of top words which represent this topic.

## E. Manual Fine-tuning

Traditional topic modeling demands extensive data cleaning, normalization, feature engineering and fine-tuning to achieve satisfactory performance. Especially when dealing with free-text survey responses which are noisy, sparse, unstructured.

In contrast, our multi agent framework leverages the inherent natural language capabilities of an LLM, significantly reducing the need for extensive data treatment and fine-tuning. LLMs exhibit remarkable robustness to variations in language, data quality, sparseness and produce consistent results.

While our framework substantially reduces human fine-tuning requirements, we acknowledge the role of prompt optimization. Achieving optimal performance necessitates careful and iterative crafting of prompts to guide the LLMs in their respective tasks.

```
You are an AI language assistant.

Yout task: A survey of tech workers was conducted to
understand the impact of AI tools in tech and given are the
responses. Enrich the free text response <column_name> with
the respondent's job title and years of experience. Add context
where ever necessary. Maintain the original sentiment and
meaning of the response. Do not introducing any new
opinions, assumptions, conclusions or extrapolations which
were not present in the original response. Keep the language
generic and standardized. Only respond with the enriched
response.

job_title: <respondent_job_title>
years_of_experience: <years_of_experience>
<column_name>: <free_text_response>

Enriched response:
```

Fig. 3. Enrichment prompt used by Agent 1

## IV. COMPARATIVE TECHNIQUES

For comparative analysis, the topics generated by MALTopic were benchmarked against those produced by LDA and BERTopic. To ensure robustness, multiple iterations of LDA and BERTopic were generated, and the iteration with the highest performance, as measured by below metrics was selected for the comparative analysis.

For the purpose of maintaining data consistency and comparability, the structured and free-text responses were integrated into a single text response via concatenation for LDA and BERTopic.

Prior to LDA topic modeling, the concatenated survey responses underwent a comprehensive preprocessing. This includes text normalization through lowercasing, removal of punctuation and extraneous whitespace, elimination of stop words, and word lemmatization.

## V. RESULTS

The full list of the topics generated from LDA, BERTopic and the MALTopic framework can be found online [16].

For illustration, topics generated for the enriched free-text field *"Concerns of AI tools in tech"* are listed below from the three techniques.

TABLE I.    MALTopic Concerns of AI tools in tech

| |
|---|
| Job Displacement Concerns - (displacement, automation, job loss, security, anxiety) |
| Skill Gaps and Training Needs - (skills, training, gap, upskill, education) |
| Data Privacy and Security Issues - (privacy, security, data, protection, risk) |
| Reliability and Trust in AI Tools - (reliability, trust, accuracy, hallucinations, skepticism) |
| Ethical Implications of AI - (ethics, bias, fairness, responsibility, implications) |
| Impact on Job Market Dynamics - (job market, hiring, availability, competition, dynamics) |
| Integration Challenges of AI Tools - (integration, challenges, workflows, adoption, resistance) |
| Cost Implications of AI Adoption - (cost, financial, budget, investment, implications) |
| Changing Role of Human Input - (human input, creativity, relevance, skills, automation) |
| Resistance to AI Adoption - (Resistance, adoption, culture, innovation, productivity) |

*Only the topic name and representative words are listed

TABLE II.  BERTopic Concerns of AI tools in tech

| |
|---|
| datum, analytic, science, experience, leadership, consulting, security, job, data, analyst |
| development, engineering, inexperienced, experience, engineer, job, blindly, trust, human, cost |
| ai, tool, experience, leadership, engineering, development, team, engineer, automation, use |
| student, inexperienced, job, interview, skill, rich, land, junior, difficult, find |
| project, program, management, experience, change, accuracy, utilise, knowledge, teach, scrape |
| product, management, quality, experience, privacy, role, value, clear, must, mitigate |

*Only representations are listed

TABLE III.  LDA Concerns of AI tools in tech

| |
|---|
| experience, leadership, ai, consulting, job, data, people, use, take, developer |
| job, inexperienced, student, ai, market, interview, analyst, take, skill, developer |
| product, management, experience, team, ai, concern, privacy, engineer, change, bias |
| engineering, development, inexperienced, change, skill, job, take, experience, people, market, |
| datum, analytic, science, tool, security, inexperienced, ai, experience, job, skill |
| management, project, program, experience, change, tool, ai, use, bias, skill |
| development, experience, engineering, ai, engineer, cost, team, use, tool, developer |

## A. Quantitative Comparison

To evaluate the performance of the three techniques, we employed a suite of metrics.

**Word Coherence:** How semantically cohesive or related the words within each topic are. Higher scores are better, and indicate that words in a topic tend to co-occur more frequently and make sense together.

$$Coherence = \frac{1}{|P|} \sum_{(w_i, w_j) \in P} PMI(w_i w_j)$$

- PMI is the pointwise mutual information for each distinct pair of works $(w_i, w_j)$.
- P is the set of all word pairs $(w_i, w_j)$.

**Word Diversity:** The proportion of unique words across all topics. Higher diversity is better, and means less overlap between the topics.

$$Diversity = \frac{|U|}{T}$$

- U is the set of unique words across all topics
- T is the total number of words across all topics

**Topic Similarity:** How similar the topics are to each other based on word embeddings. Lower value is better, and indicates more distinct topics with less semantic overlap.

$$cos(v_i, v_j) = \frac{v_i \cdot v_j}{||v_i|| \cdot ||v_j||}$$

$$Avg.\ Similarity = \frac{1}{n(n-1)} \sum_{i \neq j} cos(v_i, v_j)$$

- $v_i$ is the embedding for topic i
- $cos(v_i v_j)$ the cosine similarity between $v_i$ and $v_j$

**Document Coverage** - Percentage of documents well-represented by at least one topic. Higher coverage is better, and means the topics collectively explain more of the document collection.

$$s(d) = max_{t \in \{topics\}} cos(d, t)$$

$$Coverage = \frac{Num.\ documents\ with\ s(d) \geq \theta}{|D|}$$

- D is the document collection
- cos(d,t) is the cosine similarity between document d and topic t
- θ is a threshold parameter (set to 0.1)

Fig. 4 and 5 compare the above metrics for each topic modeling technique, and illustrate the enhanced performance of MALTopic, relative to BERTopic and LDA for the two free-text fields considered for topic modeling. Notably, MALTopic produces topics with increased word coherence and word diversity, while simultaneously reducing inter-topic similarity.

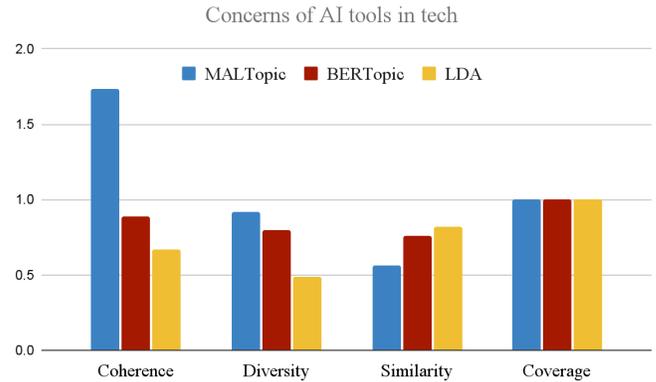

Fig. 4. Metric comparison - "Concerns of AI tools in tech"

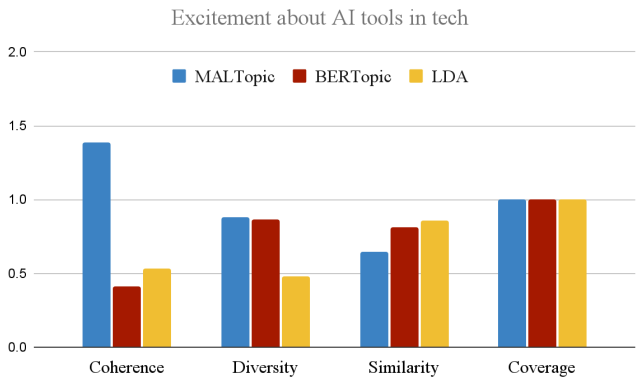

Fig. 5. Metric comparison - "Excitement about AI tools in tech"

## B. Qualitative Comparison

Table IV presents the full text content of three topics from MALTopic, selectively chosen for qualitative comparison.

TABLE IV.  MALTopic EXAMPLES

| |
|---|
| **a) Topic Name: Impact on Job Market Dynamics**<br><br>**Description** - Observations about how AI tools are changing hiring practices, job availability, and the overall landscape of employment in tech.<br>**Respondent Profile Relevance** - Notably expressed by inexperienced individuals and students, reflecting their concerns about entering a competitive job market influenced by AI.<br>**Representative words** - job market, hiring, availability, competition, dynamics |
| **c) Topic Name: Job Displacement Concerns**<br><br>**Description** - Anxiety regarding the potential loss of jobs due to the implementation of AI tools and automation.<br>**Respondent Profile Relevance** - Strongly voiced by experienced professionals in Engineering, Development, Consulting, and Data Science, reflecting their fears about job security.<br>**Representative words** - displacement, automation, job loss, security, anxiety |
| **b) Topic Name: Reliability and Trust in AI Tools**<br><br>**Description** - Skepticism about the accuracy and dependability of AI-generated outputs and the potential for errors or "hallucinations".<br>**Respondent Profile Relevance** - Strongly emphasized by experienced leaders and professionals in Engineering and Consulting, who are wary of relying on AI for critical decisions.<br>**Representative words** - reliability, trust, accuracy, hallucinations, skepticism |

Table V and VI show the closest topics generated by BERTopic and LDA, respectively. The clarity and coherence of MALTopic results highlights the framework's ability to produce human interpretable results. Notably, our framework extends beyond basic topic identification to reveal subtle underlying patterns and contextual information regarding the respondents. For example, it identified two distinct job related topics that eluded both BERTopic and LDA. The first theme centers on the challenges encountered by students and inexperienced individuals in securing employment opportunities, while the second focuses on the concerns of experienced professionals regarding potential job displacement.

TABLE V.  BERTopic EXAMPLES

| |
|---|
| **Topic Name: student_inexperienced_job_interview**<br><br>**Representations** - student, inexperienced, job, interview, skill, rich, land, junior, difficult, find. |
| **Topic Name: development_engineering_inexperienced_experience**<br><br>**Representations** - development, engineering, inexperienced, experience, engineer, job, blindly, trust, human, cost. |

TABLE VI.  LDA EXAMPLES

| |
|---|
| job, inexperienced, student, ai, market, interview, analyst, take, skill, developer |
| management, project, program, experience, change, tool, ai, use, bias, skill |

This proves our framework's ability to discern nuanced perspectives within the dataset, which is only possible due to the enhancement of free-text responses using the structured responses by Agent 1.

## VI. FUTURE WORK

A significant constraint of the proposed framework lies in the currently constrained functionality of its intelligent agents. Their role should extend beyond just augmenting free-text responses with structured data. Intelligent agents possess the potential to be leveraged for a diverse array of sophisticated tasks, such as integrating domain or industry specific context into textual information and facilitating modality translations.

Additionally, each agent should be equipped with comprehensive performance metrics and robust feedback mechanisms. Such enhancements hold the potential to improve the overall performance of the MALTopic framework and significantly reduce the occurrence of LLM hallucinations [17].

Furthermore, model bias represents a significant impediment to the generalizability of LLMs across diverse domains. An LLM trained on a corpus lacking representation from a specific domain or industry may exhibit diminished performance when applied to tasks within that domain. Consequently, an agent utilizing such biased LLM could encounter substantial challenges in executing their intended functions effectively.

The MALTopic framework, demonstrated using a single survey dataset in this research, possesses an inherent design for broad applicability across diverse textual data through the targeted modifications of agent prompts. Moreover, leveraging the inherent multimodality of the latest LLM models offers a pathway to integrate non-textual data modalities.

## VII. CONCLUSION

In this research, we addressed the two inherent limitations of traditional topic modeling algorithms specifically when applied to survey data:

- Focus primarily on analyzing textual data in isolation, and overlook the rich context provided by structured and categorical data.
- The resulting topics are often abstract and require significant human intervention for interpretation.

The MALTopic framework successfully uses LLM's superior natural language understanding to enhance free-text responses with structured response and mine human readable and interpretable topics.

## REFERENCES

[1] Blei, David M., Andrew Y. Ng, and Michael I. Jordan. "Latent dirichlet allocation." Journal of machine Learning research 3.Jan (2003): 993-1022.

[2] Grootendorst, Maarten. "BERTopic: Neural topic modeling with a class-based TF-IDF procedure." arXiv preprint arXiv:2203.05794 (2022).

[3] Sharma, Yash. "MALTopic." GitHub, 26 March 2025, https://github.com/yash91sharma/MALTopic.

[4] Kapoor, Satya, et al. "Qualitative Insights Tool (QualIT): LLM Enhanced Topic Modeling." arXiv preprint arXiv:2409.15626 (2024).

[5] Yang, Xiaohao, et al. "Llm reading tea leaves: Automatically evaluating topic models with large language models." arXiv preprint arXiv:2406.09008 (2024).

[6] Invernici, Francesco, et al. "Capturing research literature attitude towards Sustainable Development Goals: an LLM-based topic modeling approach." arXiv preprint arXiv:2411.02943 (2024).

[7] Mu, Yida, et al. "Large language models offer an alternative to the traditional approach of topic modelling." arXiv preprint arXiv:2403.16248 (2024).

[8] Al Tamime, Reham, et al. "Evaluating LLM-Generated Topics from Survey Responses: Identifying Challenges in Recruiting Participants through Crowdsourcing." 2024 IEEE Symposium on Visual Languages and Human-Centric Computing (VL/HCC). IEEE, 2024.

[9] Akash, Pritom Saha, and Kevin Chen-Chuan Chang. "Enhancing Short-Text Topic Modeling with LLM-Driven Context Expansion and Prefix-Tuned VAEs." arXiv preprint arXiv:2410.03071 (2024).

[10] Doi, Tomoki, Masaru Isonuma, and Hitomi Yanaka. "Topic modeling for short texts with large language models." Proceedings of the 62nd Annual Meeting of the Association for Computational Linguistics (Volume 4: Student Research Workshop). 2024.

[11] Chang, Shuyu, et al. "Enhanced short text modeling: Leveraging large language models for topic refinement." arXiv preprint arXiv:2403.17706 (2024).

[12] van Wanrooij, Cascha, Omendra Kumar Manhar, and Jie Yang. "Topic Modeling for Small Data using Generative LLMs."

[13] Talebirad, Yashar, and Amirhossein Nadiri. "Multi-agent collaboration: Harnessing the power of intelligent llm agents." arXiv preprint arXiv:2306.03314 (2023).

[14] OpenAI. "GPT-4o-mini." OpenAI Platform Documentation platform.openai.com/docs/models/gpt-4o-mini. Accessed March 2025.

[15] OpenAI. "openai-python." GitHub, https://github.com/openai/openai-python.

[16] Sharma, Yash. "Topics generated from MALTopic, BERTopics and LDA." GitHub, 26 March 2025, https://github.com/yash91sharma/MALTopic/blob/main/topic_modeling_results.MD.

[17] Perković, Gabrijela, Antun Drobnjak, and Ivica Botički. "Hallucinations in llms: Understanding and addressing challenges." 2024 47th MIPRO ICT and Electronics Convention (MIPRO). IEEE, 2024.